\documentclass[11pt, a4paper]{article}

\usepackage[margin=1in]{geometry}

\usepackage{mathptmx}
\usepackage[utf8]{inputenc}
\usepackage[T1]{fontenc}

\usepackage{microtype}

\usepackage{authblk}
\usepackage{amsmath, amssymb, amsfonts}
\usepackage{graphicx}
\usepackage{booktabs} % For professional looking tables
\usepackage{url}
\usepackage{tikz}
\usepackage{pgfplots}
\usepgfplotslibrary{fillbetween}
\pgfplotsset{compat=1.18}
\usetikzlibrary{arrows.meta, decorations.pathmorphing}

\usepackage{subcaption}

\newcommand{\ud}{\mathrm{d}}

\usepackage[colorlinks=true, linkcolor=blue, citecolor=blue, urlcolor=blue]{hyperref}
\usepackage{cite}

\title{\LARGE \textbf{Modeling Score Approximation Errors in Diffusion Models via Forward SPDEs}}

\author{Junsu Seo\thanks{Email: \texttt{js\_seo@snu.ac.kr}}}

\affil{Department of Mathematical Sciences, Seoul National University}

\date{}

\begin{document}
	
	\maketitle
	
	\begin{abstract}
		\noindent
		This study investigates the dynamics of Score-based Generative Models (SGMs) by treating the score estimation error as a stochastic source driving the Fokker-Planck equation. Departing from particle-centric SDE analyses, we employ an SPDE framework to model the evolution of the probability density field under stochastic drift perturbations. Under a simplified setting, we utilize this framework to interpret the robustness of generative models through the lens of geometric stability and displacement convexity. Furthermore, we introduce a candidate evaluation metric derived from the quadratic variation of the SPDE solution projected onto a radial test function. Preliminary observations suggest that this metric remains effective using only the initial 10\% of the sampling trajectory, indicating a potential for computational efficiency.
	\end{abstract}
	
	\section{Introduction}
	
	Score-based Generative Models (SGMs), also known as diffusion probabilistic models, have emerged as a dominant paradigm in generative modeling, achieving remarkable success in synthesizing high-fidelity samples across various domains \cite{ho2020denoising, song2019generative, yang2022diffusion}. Unlike prior approaches, the efficacy of SGMs is firmly rooted in the rigorous theoretical framework of Stochastic Differential Equations (SDEs), which models the data perturbation as a diffusion process and its reversal for generation \cite{song2021scorebased, anderson1982reverse}. This solid mathematical foundation has not only facilitated deep theoretical understandings, such as convergence analysis \cite{chen2024convergence} and provable robustness \cite{waibel2024scorebased}, but has also fostered extensive interdisciplinary research, bridging machine learning with diverse fields like inverse problems \cite{kawar2022denoising, chung2022diffusion} and scientific computing \cite{huang2024diffusionpde, tripura2023wdno}.
	
	In particular, the integration with Stochastic Partial Differential Equations (SPDEs) \cite{da2014stochastic, hairer2009introduction} has emerged as a natural progression within this interdisciplinary landscape, given that the SDE-based framework of SGMs intuitively extends to infinite-dimensional domains. Recent advancements have formulated Hilbert diffusion models \cite{lim2023hdm, yoon2023hdm} and infinite-dimensional score matching frameworks \cite{pidstrigach2023infinite, chen2023score}, enabling resolution-agnostic sampling. This convergence has further evolved to integrate neural operator architectures, such as Wavelet Diffusion Neural Operators \cite{tripura2023wdno}, and to accommodate non-Gaussian noise distributions via Lévy-Itô processes \cite{yoon2023levy}. Additionally, physical constraints have been incorporated to address inverse problems in fluid dynamics and other scientific domains \cite{huang2024diffusionpde}. However, these approaches mainly use SPDEs to describe how generative processes evolve stochastically in infinite-dimensional spaces.
	
	In this paper, we depart from the existing data-centric viewpoint and present an exploratory study that formulates a forward SPDE framework to analyze the \textit{score estimation error dynamics} within the Fokker-Planck equation. In this formulation, the error is not merely an additive noise in the SDE, but is modeled as a stochastic source term. This shift in perspective allows us to view the time-varying probability density not as a fixed flow, but as a random field whose evolution is dictated by the interaction between the drift and the estimation noise.
	
	Our primary motivation for this reformulation is to investigate the robustness of SGMs through the lens of geometric stability. Under a simplified setting, we explore connections between the proposed SPDE framework and concepts from optimal transport, specifically displacement convexity. We hypothesize that the robustness of generative models can be interpreted through the geometric stability of the underlying probability flow, which may dampen the impact of stochastic forcing terms. This offers a potential geometric explanation for why SGMs can generate high-quality samples even when score matching is imperfect.
	
	Furthermore, this SPDE framework introduces a new avenue for model evaluation. We derive a candidate metric, termed SPDE-Induced Evaluation Metric (SIEM), based on the quadratic variation of the SPDE solution projected onto a radial test function. Through pilot experiments utilizing CIFAR-10 and LSUN-Bedroom (downscaled to 32x32) datasets, we compared SIEM against established metrics such as FID and the 2-Wasserstein distance. These small-scale evaluations suggest that SIEM yields meaningful results, positioning it as a viable candidate for model assessment. Notably, the metric demonstrated persistent effectiveness even when computed using only the initial 10\% of the sampling trajectory. This finding implies a degree of practical utility, suggesting that SIEM could potentially facilitate efficient model monitoring.
	
	Our contributions are summarized as follows:
	\begin{itemize}
		\item First, we formulate a forward SPDE framework for SGMs by modeling the score estimation error as a stochastic source within the Fokker-Planck equation.
		\item Second, we utilize this framework to interpret the robustness of generative models, drawing connections to geometric stability and displacement convexity under simplified assumptions.
		\item Third, we introduce a candidate evaluation metric, SIEM, derived from the quadratic variation of the SPDE solution and explore its potential through preliminary comparisons with FID and the 2-Wasserstein metric on small-scale datasets.
	\end{itemize}
	\section{Notation}
	\label{sec:notation}
	
	We summarize the mathematical notations and operators used throughout this paper in Table~\ref{tab:notations}. Specifically, we explicitly define standard operators from stochastic analysis and functional calculus that appear in the derivation of the SPDE and the evaluation metric, but are not introduced in the main text.
	\begin{table}[t!]
		\caption{Summary of mathematical notations and operators.}
		\label{tab:notations}
		\begin{center}
					\renewcommand{\arraystretch}{1.2} % 가독성을 위해 행 간격 조절
					\begin{tabular}{l p{0.60\columnwidth}}
						\toprule
						\textbf{Symbol} & \textbf{Description} \\
						\midrule
						\multicolumn{2}{l}{\textit{Operators \& Calculus}} \\
						$\langle f, g \rangle$ & $L^2$ inner product, $\int f(\mathbf{x})g(\mathbf{x}) \ud\mathbf{x}$ \\
						$f \ast g$ & Spatial convolution of functions $f$ and $g$ \\
						$\frac{\delta \mathcal{E}}{\delta u}$ & Functional derivative (Variational derivative) \\
						$\nabla \cdot \mathbf{v}$ & Divergence of a vector field $\mathbf{v}$ \\
						$\Delta f$ & Laplacian operator ($\nabla \cdot \nabla f$) \\
						\midrule
						\multicolumn{2}{l}{\textit{Stochastic Analysis}} \\
						$\circ \ud \mathbf{W}$ & Stratonovich stochastic integration \\
						$\cdot \ud \mathbf{W}$ & It\^o stochastic integration \\
						$[X, X]_t$ & Quadratic variation of process $X_t$ \\
						$\mathbf{W}_Q$ & $H$-valued $Q$-Wiener process \\
						$\mathcal{L}^*$ & Fokker-Planck operator (Adjoint) \\
						\midrule
						\multicolumn{2}{l}{\textit{Functions \& Constants}} \\
						$\Gamma(z)$ & Gamma function, $\int_0^\infty x^{z-1}e^{-x} \ud x$ \\
						$\mathbb{R}^d$ & $d$-dimensional Euclidean space \\
						$L^2(\mathbb{R}^d)$ & Space of square-integrable functions \\
						\bottomrule
					\end{tabular}
		\end{center}
	\end{table}

	\section{Background}
	This section provides the foundational concepts necessary to understand our proposed framework. We explicitly distinguish between the forward diffusion process (data to noise) and the generative reverse process (noise to data), adhering to a reverse-time notation for the target trajectory. We first review the standard formulation of Score-based Generative Models (SGMs) and their connection to the Fokker-Planck Equation (FPE). We then introduce essential concepts from stochastic analysis in infinite-dimensional spaces, namely Hilbert space-valued Wiener processes, which are the mathematical tools required to formulate and analyze Stochastic Partial Differential Equations (SPDEs).
	
	\subsection{Score-based Generative Models and the Fokker-Planck Equation}
	Score-based Generative Models (SGMs) are a class of deep generative models that learn a data distribution, $p_{\text{data}}$, by reversing a predefined diffusion process \cite{song2021scorebased}. This involves two key components: a forward process and a reverse process, both described by Stochastic Differential Equations (SDEs).
	\paragraph{Forward and Reverse SDEs.}
	The \textbf{forward process} gradually perturbs data samples $\mathbf{x}_0 \sim p_{\text{data}}$ with noise over a time interval $[0, T]$. A common choice for this process is the following form of SDE:
	\begin{equation}
		d\mathbf{x}_t = f(t)\mathbf{x}_t \ud t + g(t)\ud\mathbf{w}_t, \quad t \in [0, T],
		\label{eq:bg_fwd_sde}
	\end{equation}
	where $\mathbf{w}_t$ is a standard Wiener process, and the drift and diffusion coefficients, $f(t)$ and $g(t)$, are chosen such that the distribution of $\mathbf{x}_T$, denoted $p_T$, converges to a simple, tractable prior like a standard Gaussian distribution.
	
	A key insight from stochastic calculus is that this process is reversible in time \cite{anderson1982reverse}. The corresponding \textbf{reverse-time SDE} allows for the generation of data samples by starting from the prior distribution $v_0$ (where $v_0 = p_T$) and evolving to $t=T$. Defining the time-reversed coefficients $\bar{f}_t := f(T-t)$ and $\bar{g}_t := g(T-t)$, the reverse dynamics are given by:
	\begin{equation}
		\ud\mathbf{x}_t = \left[ \bar{f}_t\mathbf{x}_t + \bar{g}_t^2 \nabla\log v_t(\mathbf{x}_t) \right] \ud t + \bar{g}_t \ud\bar{\mathbf{w}}_t,
		\label{eq:bg_rev_sde_ideal}
	\end{equation}
	where $\bar{\mathbf{w}}_t$ is a standard Wiener process. Since the true score function is unknown, SGMs train a time-dependent neural network, $\mathbf{s}_{\boldsymbol{\theta}}(t,\mathbf{x}_t)$, to approximate it. This leads to the approximated reverse SDE used for generation:
	\begin{equation}
		\ud\mathbf{x}_t = \left[ \bar{f}_t\mathbf{x}_t + \bar{g}_t^2 \mathbf{s}_{\boldsymbol{\theta}}(T-t,\mathbf{x}_t) \right] \ud t + \bar{g}_t \ud\bar{\mathbf{w}}_t.
		\label{eq:bg_rev_sde_practical}
	\end{equation}
	
	\paragraph{The Fokker-Planck Equation Perspective.}
	While SDEs describe the trajectories of individual samples, the \textbf{Fokker-Planck Equation (FPE)} provides a macroscopic view by describing the evolution of the entire probability density function. For the ideal reverse process \eqref{eq:bg_rev_sde_ideal}, the density $v_t$ evolves according to its corresponding FPE. Similarly, the density generated by the practical process \eqref{eq:bg_rev_sde_practical} starting from $u_0=v_0=p_T$, which we denote $u_{t}$, follows a slightly different FPE.
	
	By comparing the FPE for the practical process to that of the ideal one, we can isolate the effect of the score approximation error. The dynamics of $u_{t}$ can be written as:
	\begin{equation}
		\partial_t u_{t} = \mathcal{L}^* u_{t} - \nabla \cdot \left( \bar{g}_t^2 u_{t} \left[ \mathbf{s}_{\boldsymbol{\theta}} - \nabla \log v_t \right] \right),
		\label{eq:bg_fpe_with_error}
	\end{equation}	
	where $\mathcal{L}^*$ is the operator for the ideal reverse dynamics. The discrepancy term $- \nabla \cdot \left( \bar{g}_t^2 u_{t} \left[ \mathbf{s}_{\boldsymbol{\theta}} - \nabla \log v_t \right] \right)$ drives $u_t$ away from the target $v_t$.	
	
	\subsection{Stochastic Analysis in Infinite Dimensions}
	\label{subsec: Stochastic Analysis in Infinite Dimensions}
	To rigoroulsy describe the evolution of probability densities, we extend stochastic calculus to the Hilbert space $H = L^2(\mathbb{R}^d)$. We model the noise as an $H$-valued $Q$-Wiener process. Specifically, to account for the vector-valued score function, we consider $\mathbf{W}_Q = (W^1_{Q_1}, \dots, W^d_{Q_d})$ where each component is a scalar $Q$-Wiener process in $L^2(\mathbb{R}^d)$. This allows the noise to interact properly with the divergence operator in the Fokker-Planck setting.
	
	\paragraph{Conservative Stochastic Dynamics.}
	Unlike standard SDEs for particles, the density evolution must conserve total mass. We model this by introducing stochasticity in the flux term. Formally, the stochastic forcing is modeled as a conservative divergence term $\nabla \cdot (u_t \circ \mathrm{d}\mathbf{W}_Q)$, ensuring that noise redistributes probability mass without creating or destroying it.
	
	\paragraph{Choice of Integration Scheme.}
	A critical distinction in this work is the choice of stochastic integration. We employ two interpretations depending on the analytical context:
	\begin{itemize}
		\setlength\itemsep{0em}
		\item \textbf{Stratonovich Integral ($\circ$):} In Sections \ref{sec:SFPE_motivation} and \ref{sec_simplified_model}, we adopt the Stratonovich interpretation. Unlike It\^o calculus, Stratonovich integration satisfies the standard chain rule of ordinary calculus. This property is essential for preserving the geometric structure of the density flow and analyzing energy dissipation without the additional correction terms required by It\^o's Lemma.
		\item \textbf{It\^o Integral ($\cdot$):} In Section 5, we utilize the It\^o interpretation for model evaluation. The It\^o integral possesses the martingale property (i.e., zero expectation given the present), which is crucial for defining an unbiased estimator for the quadratic variation metric.
	\end{itemize}
	We implicitly assume the standard conversion between these forms, which introduces a drift correction term (Wong-Zakai correction), accounted for in our drift definitions where appropriate.
	
	\section{Motivation and Formulation of the Forward SPDE}
	\label{sec:SFPE_motivation} 
	We can describe the evolution of a distribution $u_{t}$ generated using a learned score $\mathbf{s}_{\boldsymbol{\theta}}$ with the reverse FPE:
	\begin{align}
		\partial_t u_{t} &= \underbrace{-\nabla\cdot(u_{t}(\bar{f}_{t}\mathbf{x}+\bar{g}^2_t \nabla\log v_t))+\frac{\bar{g}^2_t}{2}\Delta u_{t}}_{\text{Ideal FPE dynamics}}-\underbrace{\bar{g}^2_t\nabla\cdot\left(u_{t}\left(\bar{\mathbf{s}}_{\boldsymbol{\theta}}-\nabla\log v_t\right)\right)}_{\text{Drift perturbation } D(t,x)}. \label{eq:ideal_reverse_FP_appendix}
	\end{align}
	where $\bar{f}_{t}:=f(T-t)$, $\bar{\mathbf{s}}_{\boldsymbol{\theta}}:=\mathbf{s}_{\boldsymbol{\theta}}(T-t,x)$ and $\bar{g}_t:=g(T-t)$.
	
	The first term in \eqref{eq:ideal_reverse_FP_appendix} represents the ideal dynamics that would occur if drift perturbation $D$ is zero. In this case, the trajectory should be $v_t$ so that we can trace the probability distribution the trajectory exactly . However, the presence of the drift perturbation, which stems from the discrepancy between the learned score $\mathbf{s}_{\boldsymbol{\theta}}$ and the true score $\nabla\log v_t$, fundamentally alters the dynamics. This term acts as a complex, state-dependent perturbation, making the evolution of $u_{t}$ deviate from the ideal path and rendering the full PDE in \eqref{eq:ideal_reverse_FP_appendix} analytically intractable.
	\begin{figure}[t!]
		\centering
		\begin{tikzpicture}
			\begin{axis}[
				width=0.95\linewidth,
				height=6cm,
				axis lines=middle,
				xlabel={Time (Reverse Process)},
				ylabel={State Space $\mathbf{x}$},
				xmin=0, xmax=6.5,
				ymin=-2.5, ymax=2.5,
				xtick=\empty, ytick=\empty,
				% 축 레이블 위치 조정
				xlabel style={at={(axis description cs:1.0,-0.05)},anchor=north east},
				ylabel style={at={(axis description cs:-0.05,0.5)},rotate=90,anchor=south},
				title={\textbf{Robustness via Geometric Contractivity}},
				legend style={at={(0.95,0.95)}, anchor=north east, draw=none, fill=white, fill opacity=0.8},
				clip=false
				]
				
				% 1. Stability Funnel (Contractivity Zone)
				% [수정] 범례 순서를 차지하지 않도록 'forget plot' 추가
				\addplot[name path=upper, draw=none, domain=0:6, samples=50, forget plot] {2.0*exp(-0.3*x)};
				\addplot[name path=lower, draw=none, domain=0:6, samples=50, forget plot] {-2.0*exp(-0.3*x)};
				\addplot[gray!10, opacity=0.8, forget plot] fill between [of=upper and lower];
				\node[gray!60, font=\scriptsize, rotate=-10] at (axis cs:3.5, 0.8) {Contractivity Zone};
				
				% 2. Vector Field (Restoring Force)
				% [수정] 범례 순서를 차지하지 않도록 'forget plot' 추가
				\addplot[
				blue!30,
				quiver={
					u=1,
					v=-0.8*y,
					scale arrows=0.2,
				},
				-stealth,
				samples=15,
				samples y=10,
				domain=0:6,
				domain y=-2.2:2.2,
				forget plot % <--- 중요: 화살표가 범례에 나오지 않게 함
				] {0};
				
				% 3. Ideal Trajectory (v_t)
				% [수정] 그래프 바로 뒤에 addlegendentry를 붙여서 스타일 자동 적용
				\addplot[
				dashed,
				black!80,
				line width=1.2pt,
				domain=0:6.2
				] {0};
				\addlegendentry{Ideal Path $v_t$}
				
				% 4. Stochastic Trajectory (u_t)
				% [수정] 그래프 바로 뒤에 addlegendentry를 붙여서 스타일 자동 적용
				\addplot[
				color=orange!90!red,
				line width=1.0pt,
				rounded corners=1pt, 
				] coordinates {
					(0.0, 1.5) (0.2, 1.7) (0.4, 1.2) (0.6, 1.4) (0.8, 0.8) 
					(1.0, 1.1) (1.2, 0.5) (1.4, -0.2) (1.6, 0.2) (1.8, -0.5) 
					(2.0, -0.1) (2.2, -0.8) (2.4, -0.3) (2.6, -0.9) (2.8, -0.4) 
					(3.0, -0.1) (3.2, 0.3) (3.4, -0.2) (3.6, 0.4) (3.8, 0.1) 
					(4.0, -0.3) (4.2, 0.0) (4.4, -0.2) (4.6, 0.15) (4.8, -0.1)
					(5.0, 0.05) (5.2, -0.1) (5.4, 0.02) (5.6, -0.05) (5.8, 0.02) (6.0, 0.0)
				};
				\addlegendentry{Perturbed Path $u_t$}
				
				% 5. Annotations (설명)
				\node[blue!50!black, font=\footnotesize, align=center] at (axis cs:1.5, -1.8) {\textbf{Restoring Force}\\(Geometry)};
				\draw[->, blue!50!black, thick] (axis cs:1.5, -1.5) -- (axis cs:1.5, -0.8);
				
				\node[orange!90!red, font=\footnotesize, align=center] at (axis cs:1.2, 2.3) {\textbf{Stochastic}\\ \textbf{Perturbation}};
				\draw[->, orange!90!red, dashed, thick] (axis cs:1.2, 1.9) -- (axis cs:1.0, 1.3);
				
				\node[circle, fill=black, inner sep=1.5pt, label={above right:$v_T$ (Target)}] at (axis cs:6.0, 0) {};
				
				% [삭제됨] 하단에 있던 수동 addlegendimage 코드는 삭제했습니다.
				% 위에서 \addlegendentry를 직접 연결했으므로 자동으로 생성됩니다.
				
			\end{axis}
		\end{tikzpicture}
		\caption{Visualization of robustness in the simplified model. The \textbf{gray funnel} and \textbf{blue vector field} illustrate the geometric contractivity, which acts as a restoring force towards the ideal path. Although the trajectory $u_t$ (orange) is continuously subjected to stochastic perturbations ("shaking"), the geometric stability confines the error, ensuring convergence to the target $v_T$.}
		\label{fig:robustness_viz}
	\end{figure}
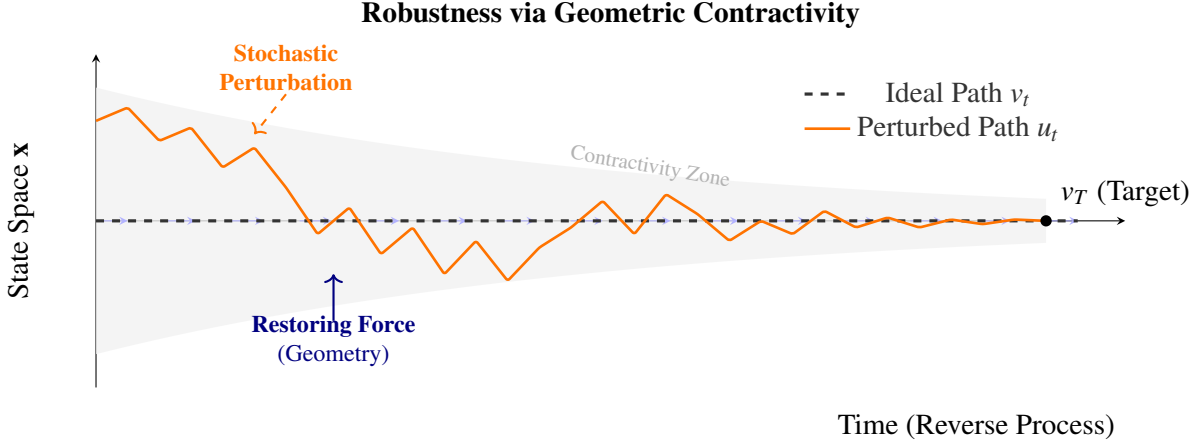		
	We model the discrepancy $\mathbf{s}_{\boldsymbol{\theta}}-\nabla\log v_t$ as a sample path of $\boldsymbol{\mu}_t + \mathbf{W}_Q$. Here, $\boldsymbol{\mu}_t \equiv \boldsymbol{\mu}(t, \mathbf{x})$ denotes the deterministic drift vector field representing the systematic bias, and $\mathbf{W}_Q$ represents the vector-valued Wiener process defined in Section \ref{subsec: Stochastic Analysis in Infinite Dimensions}. We motivate this stochastic treatment by observing the training methodology of the score-matching loss. The uniform random sampling of the time variable $t$ suggests that the error manifests with stochastic characteristics rather than a purely deterministic structure. Proceeding with this intuition, we replace the deterministic drift perturbation with this stochastic model.
	
    we define the effective drift vector $\mathbf{h}(t, \mathbf{x})$ as:
	\begin{equation}
		\mathbf{h}(t, \mathbf{x}) := \bar{f}_{t}\mathbf{x}+\bar{g}^{2}_{t} (\nabla\log v_t+\boldsymbol{\mu}_t).
		\label{eq:eff_drift}
	\end{equation}
	By replacing the deterministic drift perturbation $D(t,x)$ with this stochastic model, we transition from a standard PDE to a stochastic partial differential equation (SPDE):
	\begin{align}
		\ud u_{t} &= \left[ -\nabla\cdot(u_{t}\mathbf{h}(t, \mathbf{x}))+\frac{\bar{g}^{2}_{t}}{2}\Delta u_{t} \right] \ud t -\bar{g}^{2}_{t}\nabla\cdot(u_{t}\circ \ud\mathbf{W}_Q ) \label{eq_MainSPDEs}
	\end{align}
	The final term is a \textbf{conservative noise} that ensures the total mass of the solution is conserved over time. This allows the solution to be interpreted as a path-dependent probability distribution.
	\section{Analysis of a Simplified Model for Robustness}
	\label{sec_simplified_model}
	Empirical observations suggest that score-based models $\mathbf{s}_{\boldsymbol{\theta}}$ can generate high-quality samples even when training is truncated at an early stage. While several interpretations of this phenomenon have been proposed~\cite{mimikosstamatopoulos2024scorebasedgenerativemodelsprovably,chen2023sampling,chen2023lowdim,koehler2023statistical,debortoli2022convergence}, we present a novel perspective using Eq.~\eqref{eq_MainSPDEs} as a proof of concept. Our key insight is that the trajectory defined by the ideal FPE dynamics is sufficiently robust that stochastic perturbations cannot significantly cause deviations.
	
	Specifically, we consider a parameterization where $\bar{f}_{t}=0$, $\bar{g}_{t}=\sqrt{2}$, and $\boldsymbol{\mu}_t=0$. Under these conditions, the target distribution evolves as $v_t = k_{T-t} \ast p_0$ (reverse-time heat equation). Since the SDEs in score-based models are intrinsically rooted in such diffusive processes, this simplified model offers transparent insights.
	
	In this setting, the dynamics of the generated distribution $u_t$ are governed by the following Stratonovich SPDE:
	\begin{align}
		\ud u_t = \underbrace{\nabla\cdot\left(u_t\nabla \log \frac{u_t}{v_t} \right) \ud t}_{\text{Wasserstein Gradient Flow}} \underbrace{-\nabla\cdot(u_t\circ \mathrm{d}\mathbf{W}_Q )}_{\text{Stochastic Perturbation}}. \label{eq_anal_1}
	\end{align}
	Here, we have reformulated the ideal dynamics to explicitly highlight their geometric structure: the first term represents the gradient flow of the free energy $\mathcal{E}_t[u] = \text{KL}(u \| v_t)$ in the 2-Wasserstein space. This geometric structure acts as a restoring force, continually guiding $u_t$ towards the target trajectory $v_t$.
	
	\subsection{Geometric Stability via Displacement Convexity}
	
	To analyze the stability physically, we consider the time evolution of the free energy $\mathcal{E}_t$. A theoretical challenge is that standard geometric inequalities typically assume a static potential, whereas our target potential $\Phi_t = -\log v_t$ is time-dependent. To address this while retaining physical intuition, we employ a \textit{quasi-static approximation}. We partition the time interval $[0, T]$ into $N$ sub-intervals $[t_i, t_{i+1}]$. Within each short interval, we freeze the potential as $\Phi \approx -\log v_{\bar{t}_i}$ (where $\bar{t}_i$ is the midpoint), treating the system as relaxing towards a momentary equilibrium state.
	
	We invoke the concept of \textit{Displacement Convexity} from optimal transport. Since $v_t$ arises from diffusion on a compact domain, it tends to become log-concave~\cite{LeeVazquez2003}, implying that the potential $\Phi$ is convex. This convexity ensures that the energy landscape is "steep" enough to force rapid convergence.
	
	Now, we formally derive the time evolution of the energy $\mathcal{E}[u_t]$ to observe the competition between restoration and noise. For brevity, let $\psi \triangleq \log(u/v_{\bar{t}_i})$ denote the relative potential. Taking the time derivative and applying the chain rule heuristically to Eq.~\ref{eq_anal_1}, we obtain:
	\begin{align}
		\frac{\mathrm{d}}{\mathrm{d}t}\mathcal{E}[u_t] &= \int \frac{\delta \mathcal{E}}{\delta u} (\partial_t u) \mathrm{d}\mathbf{x} \nonumber \\
		&= -\underbrace{\int u \left\| \nabla \psi \right\|^2 \mathrm{d}\mathbf{x}}_{\text{Dissipation}} + \underbrace{\int \nabla \psi \cdot (u \circ \ud\mathbf{W}_Q) \,\mathrm{d}\mathbf{x}}_{\text{Noise Power}}.
	\end{align}
	The first term is the energy dissipation rate due to the gradient flow. Under the assumption of displacement convexity, the Log-Sobolev inequality (LSI) guarantees that this dissipation dominates the energy itself, i.e., $\text{Dissipation} \ge 2\lambda \mathcal{E}[u_t]$ for some $\lambda > 0$. The second term represents the effective power injected by the stochastic perturbation. Modeling this noise interaction as bounded by a constant $\eta$, we arrive at the following differential inequality:
	\begin{align}
		\frac{\mathrm{d}}{\mathrm{d}t}\mathcal{E}[u_t] \le -2\lambda \mathcal{E}[u_t] + \eta. \label{eq_diff_ineq}
	\end{align}
	This inequality reveals the physical mechanism of robustness: the geometric restoration rate $2\lambda$ exponentially suppresses the error, countering the constant noise injection $\eta$. Consequently, the deviation $\mathcal{E}$ does not diverge but stabilizes within a bounded region (see Figure~\ref{fig:robustness_viz}), explaining why the generated samples remain close to the target distribution despite early truncation.
	
	\section{SPDE-Induced Evaluation Metric}
	This section introduces a practical evaluation metric derived from the SPDE presented earlier. Recalling the effective drift $\mathbf{h}(t, \mathbf{x})$ from Eq.~\eqref{eq:eff_drift}, we restate the equation:
	\begin{align}
		\ud u_t &= \left[-\nabla\cdot(u_t\mathbf{h}(t, \mathbf{x}))+\frac{\bar{g}^{2}_{t} }{2}\Delta u_t\right]\ud t -\bar{g}^{2}_{t}\nabla\cdot(u_t\circ \ud\mathbf{W}_Q ).
	\end{align}
	Projecting this equation onto a test function $\phi$, we obtain:
	\begin{align}
		\ud\langle u_t, \phi \rangle &= \left\langle -\nabla\cdot\left(u_t\mathbf{h}(t, \mathbf{x})\right) + \frac{\bar{g}^{2}_{t}}{2}\Delta u_t, \, \phi \right\rangle \ud t - \langle \bar{g}^{2}_{t}\nabla\cdot (u_t\circ \ud\mathbf{W}_Q), \phi \rangle 
	\end{align}
	We note that the quadratic variation of the projected process is identical under both Stratonovich and It\^o interpretations, which justifies analyzing the noise term via It\^o calculus to leverage its martingale property\cite{karatzas1991brownian}. Let $X_t := \langle u_t, \phi \rangle$. Its martingale part, $M_t$, is the Itô integral component of the stochastic term, $-\langle \bar{g}^{2}_{t}\nabla\cdot(u_t \ud\mathbf{W}_Q ),\phi\rangle$. The quadratic variation $[X,X]_t$ is determined solely by this martingale part, $[M,M]_t$. We approximate it by summing the squared increments of $M_t$ over a small time interval $\Delta t$. Each increment $\Delta M_t$ is driven by the noise increment $\Delta W_Q(x) \approx \sqrt{\Delta t} Z_t(x)$, where $Z_t(x)$ is a realization of spatial noise. Crucially, in our score-based model, this noise realization $Z_t(x)$ is given by $\mathbf{s}_{\boldsymbol{\theta}}-\nabla\log v_t-\boldsymbol{\mu}_t$. Here, $\boldsymbol{\mu}_t$ is the systematic drift field defined in Eq. \eqref{eq:eff_drift}.
	
	Thus, we can approximate $[X,X]_t$ as:
	\begin{align*}
		[X,X]_t &\approx \sum (\Delta M_t)^2 \approx \sum \left(\langle \bar{g}^{2}_{t}\nabla\cdot(u_t Z_t ),\phi\rangle\right)^2\Delta t\\
		&\approx \sum \left( \bar{g}^{2}_{t} \langle u_t (\mathbf{s}_{\boldsymbol{\theta}}-\nabla\log v_t-\boldsymbol{\mu}_t), \nabla\phi \rangle \right)^2\Delta t \\
		&=\sum \left(  \bar{g}^{2}_{t}\langle u_t (\mathbf{s}_{\boldsymbol{\theta}}-\nabla\log v_t),\nabla\phi\rangle-\boldsymbol{\mu}_{t,\phi}\right)^2\Delta t,
	\end{align*}
	where $\mu_{t,\phi} := \bar{g}^{2}_{t}\langle u_t \boldsymbol{\mu}_t, \nabla\phi \rangle$.
	
	We employ the $d$-dimensional radial Lorentz function as our test function $\phi(\mathbf{x})$, defined as:
	$$\phi(\mathbf{x}) = \frac{\Gamma\left(\frac{1+d}{2}\right)}{\pi^{\frac{d+1}{2}} \epsilon^{\frac{d}{2}}} \left( 1 + \frac{\|\mathbf{x}\|^2}{\epsilon} \right)^{-\frac{d+1}{2}}$$
	
	Using the property of $\nabla\phi(\mathbf{x}) = -(d+1)\frac{\mathbf{x}}{\epsilon+\|\mathbf{x}\|^2}\phi(\mathbf{x})$, we rewrite the inner product term as:
	\begin{align*}
		\bar{g}^{2}_{t}\langle u_t (\mathbf{s}_{\boldsymbol{\theta}}-\nabla\log v_t),\nabla\phi\rangle= -(d+1)\bar{g}^{2}_{t} \left\langle u_t (\mathbf{s}_{\boldsymbol{\theta}}-\nabla\log v_t)  , \frac{\mathbf{x}}{\epsilon+\|\mathbf{x}\|^2}\phi \right\rangle    
	\end{align*}
	
	In high-dimensional settings, the concentration of measure phenomenon causes the radial Lorentz function $\phi(\mathbf{x})$ to assume nearly constant values over the typical set of $\mathbf{x}$. Consequently, $\phi(\mathbf{x})$ acts effectively as a scalar multiplier rather than a variable weighting function within the inner product. This stability allows us to approximate the term by factoring out $\phi(\mathbf{x})$, leading to the following simplification:
	\begin{align*}
		\bar{g}^{2}_{t}\langle u_t (\mathbf{s}_{\boldsymbol{\theta}}-\nabla\log v_t),\nabla\phi\rangle&= -(d+1)\bar{g}^{2}_{t} \left\langle u_t (\mathbf{s}_{\boldsymbol{\theta}}-\nabla\log v_t)  , \frac{\mathbf{x}}{\epsilon+\|\mathbf{x}\|^2}\phi \right\rangle \\
		&\approx -(d+1)\bar{g}^{2}_{t} C_{\phi} \left\langle u_t (\mathbf{s}_{\boldsymbol{\theta}}-\nabla\log v_t)  , \frac{\mathbf{x}}{\epsilon+\|\mathbf{x}\|^2} \right\rangle
	\end{align*}
	
	We normalize the coefficient $(d+1)C_\phi$ to 1 to preclude notational ambiguity with differentials, as this constant does not alter the metric's relative structure. However, in numerical experiments, the magnitude of the vector term $\mathbf{x}/(\epsilon+\|\mathbf{x}\|^2)$ diminishes rapidly as the dimension $d$ increases. To counteract this attenuation and ensure numerical stability, we empirically scale the term by the dimension $d$. Interpreting the inner product as an expectation with respect to the distribution $u_t$, we obtain the following Monte Carlo approximation:
	\begin{align*}\mathbb{E}_{\mathbf{x}_t \sim u_t} \bigg[ \bar{g}_t^2 \left( \mathbf{s}_{\boldsymbol{\theta}}(\mathbf{x}_t, t) - \nabla\log v_t(\mathbf{x}_t) \right) \cdot \frac{\mathbf{x}_t}{\epsilon+\|\mathbf{x}_t\|^2} \bigg].
	\end{align*}
	Direct estimation via sampling from $u_t$ is intractable because the integrand depends on the score discrepancy relative to the marginal density $v_t$.To resolve this, we employ importance sampling to shift the expectation to the tractable distribution $v_t$, yielding the following computable expression $\xi(t)$:
	\begin{align}
		\mathbb{E}_{\mathbf{x}_t\sim v_t}\bigg[ \bar{g}_t^2 \frac{u_t}{v_t} (\mathbf{s}_{\boldsymbol{\theta}}(\mathbf{x}_t,t)-\nabla\log v_t(\mathbf{x}_t))\cdot\frac{\mathbf{x}_t}{\epsilon+\|\mathbf{x}_t\|^2} \bigg]. \label{eq:drift_estimator}
	\end{align}
	With the exception of the density ratio $u_t/v_t$, all terms in Equation \eqref{eq:drift_estimator} are directly computable via sampling (see Appendix A for the detailed derivation). We estimate the ratio $u_t/v_t$ using the approximation method described in Appendix B.
	
	For practical application, we refine the underlying quadratic variation. Since the quadratic variation accumulates squared fluctuations, it inherently scales as the second moment of the process. To recover a metric that is linearly proportional to the magnitude of the discrepancies—analogous to deriving standard deviation from variance—we apply a square root. 
	
	The final form of our metric, which we call the \textbf{SPDE-induced Evaluation Metric}, is given by:
	\begin{align}
		\text{SIEM} = \sqrt{\sum \bigg( \xi(t) -\mu_{t,\phi} \bigg)^2 \Delta t}\label{SIEM}.
	\end{align}
	The term $\mu_{t,\phi}$ represents the systematic bias of the score discrepancy. Since direct computation of this term is intractable, we approximate it by exploiting the separation of time scales between the global bias and the local stochastic fluctuations. Under the premise that the systematic bias exhibits temporal regularity distinct from the high-frequency oscillations of $\xi_t$, we estimate $\mu_{t,\phi}$ by applying a temporal Gaussian smoothing kernel to the trajectory of $\xi_t$.
	
	\section{Experiments}
	\begin{figure}[t!] % Changed from [!h] to [ht]
		\centering
		\includegraphics[width=0.95\linewidth]{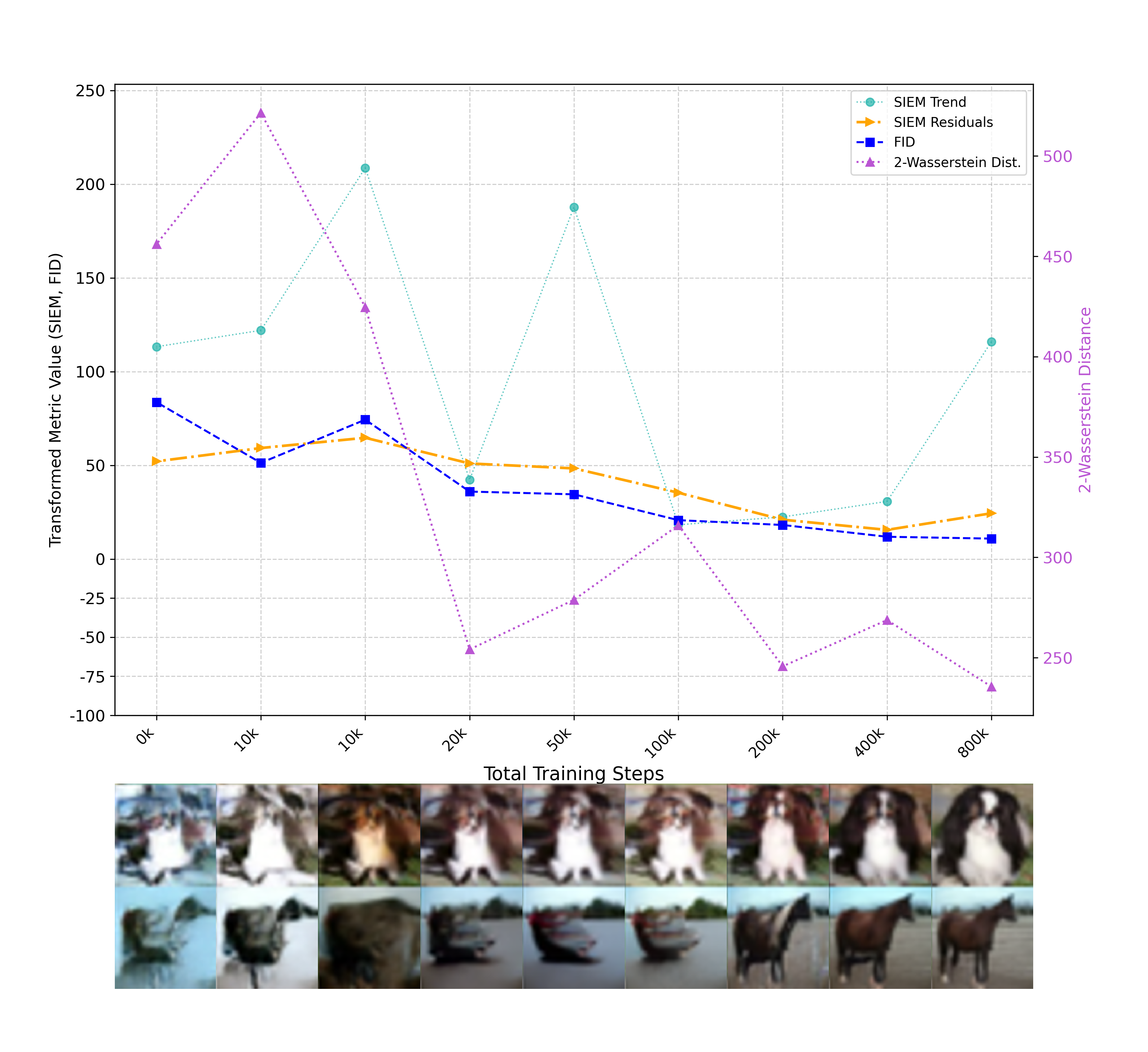}
		\caption{Evolution of $\boldsymbol{\mu}_t$ (light green), SIEM (orange), FID (blue), and 2-Wasserstein distance (light purple) during U-Net training on CIFAR-10. Generated samples at various training steps are shown at the bottom.}
		\label{fig:training_dynamics_cifar10}
	\end{figure}
	\label{chap:experiments}
	We conduct an experiment to validate our SPDE-Induced Evaluation Metric (SIEM). Specifically, we investigate whether SIEM can effectively track the performance improvement of a generative model during training by analyzing its correlation with standard metrics like FID and the 2-Wasserstein distance.
	
	\subsection{Experimental Setup}
	\label{sec:setup}
	For comprehensive details regarding datasets, model architecture, and training configurations, please refer to Appendix \ref{sec:appendix_experimental_details}.
	
	\paragraph{Models and Datasets}
	Our experiment is based on the Denoising Diffusion Probabilistic Model (DDPM) framework \cite{ho2020denoising} with 1000 diffusion timesteps. We use a standard U-Net architecture \cite{ronneberger2015u-net} as the noise prediction network. The model is trained on two datasets: CIFAR-10 \cite{krizhevsky2009learning} and a 32x32 resized version of LSUN-Bedrooms \cite{yu2015lsun}. All images were normalized to $[-1, 1]$.
	
	\paragraph{Evaluation Metrics}
	We benchmark our proposed metric, SIEM, against two established metrics: Fr$\acute{\text{e}}$chet Inception Distance (FID) \cite{heusel2017gans} for perceptual quality and the 2-Wasserstein distance \cite{feydy2019interpolating}, chosen to reflect the theoretical framework presented in Section \ref{sec_simplified_model}.
	
	\subsection{Correlation with Training Progress}
	\label{sec:exp_training_dynamics}
	To assess the validity of SIEM, we trained a DDPM with a U-Net backbone on both CIFAR-10 and LSUN-Bedrooms for approximately 800,000 iterations. We periodically evaluated model checkpoints to observe the evolution of metrics as the model improved. To account for the longer training duration required for significant performance gains in later stages, we performed measurements at exponentially increasing intervals, doubling from 3,125 to 800,000 iterations.
	
	\paragraph{Results}
	Our results demonstrate that SIEM correlates strongly with standard metrics, particularly FID, indicating that SIEM effectively tracks the improvement in sample quality throughout training. As shown in Figure~\ref{fig:training_dynamics_cifar10} for CIFAR-10, SIEM consistently decreases, mirroring the trend of the FID score and aligning with the enhanced visual quality of generated samples. The Spearman correlation between SIEM and FID is 0.8898, which is significantly higher than that with the 2-Wasserstein distance (0.5094), as detailed in Table~\ref{tab:correlation_matrix}. In contrast, the intermediate metric $\boldsymbol{\mu}_t$ exhibited a weaker correlation with FID compared to SIEM and did not show a statistically significant correlation ($p > 0.05$) with the 2-Wasserstein distance.
	
	A key observation is that this high correlation is maintained even when SIEM is calculated using only a subset of timesteps (e.g., from 900 to 999). For instance, the correlation between SIEM and FID remains high at 0.8605 when using this reduced range. This suggests that SIEM can provide a reliable evaluation with a significantly reduced computational budget compared to metrics requiring full sample generation.
	
\begin{table}[!h]
    \centering
    \caption{Comparisons of Spearman Correlation Matrices and P-values across different DDPM timestep intervals. The \textbf{top row} represents the full diffusion trajectory (Timesteps 0-999), while the \textbf{bottom row} focuses on the initial denoising stages (timesteps 900-999). Results are averaged over 5 runs for each of the 36 model configurations (comprising 18 CIFAR-10 and 18 LSUN models, varied by 2 initialization schemes and 9 training durations).}
    \label{tab:correlation_matrix}
    
    \vspace{0.3cm} % 캡션과 표 사이의 간격 조정 (필요 시 조절)

    % =======================================================
    % TOP ROW: Timesteps 0-999
    % =======================================================
    \begin{minipage}{0.48\textwidth}
        \centering
        \textbf{Timesteps 0-999: Correlation Coefficients} \\[0.1cm]
        \resizebox{\linewidth}{!}{%
            \begin{tabular}{@{}lcccc@{}}
                \toprule
                & $\|\boldsymbol{\mu}_t\|_2$ & \textbf{SIEM} & \textbf{FID} & \textbf{2-Wasser} \\
                \midrule
                $\|\boldsymbol{\mu}_t\|_2$ & 1.0000 & 0.6342 & 0.5014 & 0.0831 \\
                \textbf{SIEM}              & 0.6342 & 1.0000 & 0.8898 & 0.5094 \\
                \textbf{FID}               & 0.5014 & 0.8898 & 1.0000 & 0.5992 \\
                \textbf{2-Wasser}          & 0.0831 & 0.5094 & 0.5992 & 1.0000 \\
                \bottomrule
            \end{tabular}%
        }
    \end{minipage}
    \hfill
    \begin{minipage}{0.48\textwidth}
        \centering
        \textbf{Timesteps 0-999: P-values} \\[0.1cm]
        \resizebox{\linewidth}{!}{%
            \begin{tabular}{@{}lcccc@{}}
                \toprule
                & $\|\boldsymbol{\mu}_t\|_2$ & \textbf{SIEM} & \textbf{FID} & \textbf{2-Wasser} \\
                \midrule
                $\|\boldsymbol{\mu}_t\|_2$ & 0.0000 & 0.0000 & 0.0018 & 0.6298 \\
                \textbf{SIEM}              & 0.0000 & 0.0000 & 0.0000 & 0.0015 \\
                \textbf{FID}               & 0.0018 & 0.0000 & 0.0000 & 0.0001 \\
                \textbf{2-Wasser}          & 0.6298 & 0.0015 & 0.0001 & 0.0000 \\
                \bottomrule
            \end{tabular}%
        }
    \end{minipage}

    \par\bigskip % 행 간 간격

    % =======================================================
    % BOTTOM ROW: Timesteps 900-999
    % =======================================================
    \begin{minipage}{0.48\textwidth}
        \centering
        \textbf{Timesteps 900-999: Correlation Coefficients} \\[0.1cm]
        \resizebox{\linewidth}{!}{%
            \begin{tabular}{@{}lcccc@{}}
                \toprule
                & $\|\boldsymbol{\mu}_t\|_2$ & \textbf{SIEM} & \textbf{FID} & \textbf{2-Wasser} \\
                \midrule
                $\|\boldsymbol{\mu}_t\|_2$ & 1.0000 & 0.7030 & 0.5176 & 0.3441 \\
                \textbf{SIEM}              & 0.7030 & 1.0000 & 0.8605 & 0.5707 \\
                \textbf{FID}               & 0.5176 & 0.8605 & 1.0000 & 0.5992 \\
                \textbf{2-Wasser}          & 0.3441 & 0.5707 & 0.5992 & 1.0000 \\
                \bottomrule
            \end{tabular}%
        }
    \end{minipage}
    \hfill
    \begin{minipage}{0.48\textwidth}
        \centering
        \textbf{Timesteps 900-999: P-values} \\[0.1cm]
        \resizebox{\linewidth}{!}{%
            \begin{tabular}{@{}lcccc@{}}
                \toprule
                & $\|\boldsymbol{\mu}_t\|_2$ & \textbf{SIEM} & \textbf{FID} & \textbf{2-Wasser} \\
                \midrule
                $\|\boldsymbol{\mu}_t\|_2$ & 0.0000 & 0.0000 & 0.0012 & 0.0399 \\
                \textbf{SIEM}              & 0.0000 & 0.0000 & 0.0000 & 0.0003 \\
                \textbf{FID}               & 0.0012 & 0.0000 & 0.0000 & 0.0001 \\
                \textbf{2-Wasser}          & 0.0399 & 0.0003 & 0.0001 & 0.0000 \\
                \bottomrule
            \end{tabular}%
        }
    \end{minipage}
\end{table}

	\section{Discussion}
	\label{sec:discussion}
	
	The experimental results presented in Section~\ref{chap:experiments} provide empirical support for the SPDE-based framework proposed in this study. In this section, we interpret these findings in the context of our theoretical assumptions, justify our methodological choices, and discuss the limitations of the current modeling.
	
	\textbf{Geometric Validity and Wasserstein Connections.}
	A key premise of our framework is that the robustness of generative models can be analyzed through the geometry of the probability flow. The observed correlation between SIEM and the 2-Wasserstein distance ($W_2$), while distinct from that of FID, validates this assumption. It suggests that the error dynamics are not merely additive noise but interact with the transport geometry in a way that is captured by Wasserstein metrics. This reinforces the utility of viewing score estimation errors through the lens of optimal transport.
	
	\textbf{SIEM as a Measure of Structural Volatility.}
	Our analysis highlights a critical distinction between the mean drift error $\|\boldsymbol{\mu}_t\|_2$ and SIEM. While $\|\boldsymbol{\mu}_t\|_2$ naturally correlates with the score-matching objective—decreasing as the model converges—SIEM exhibits a consistently higher correlation with perceptual quality (FID). We interpret $\|\boldsymbol{\mu}_t\|_2$ as the magnitude of the systematic bias, whereas SIEM, derived from the quadratic variation, captures the \textit{residual volatility} of the trajectory. The finding that SIEM remains highly correlated with FID even when computed on truncated trajectories (e.g., the first 10\% of sampling) suggests that the structural instability of the generative process is detectable early in the reverse diffusion. This indicates that SIEM measures a fundamental geometric property of the error that is distinct from, and potentially more diagnostic than, the raw error magnitude.
	
	\textbf{Limitations.}
	As an exploratory study, this work has limitations. Primarily, we modeled the score error as a generic stochastic source without characterizing its specific spatial structure. In reality, the error induced by neural networks likely concentrates on specific eigenbases. The inability to experimentally identify the eigenbasis of the stochastic term $\ud\mathbf{W}_Q$ and the precise form of the drift $\boldsymbol{\mu}(t,x)$ is a limitation. A deeper understanding of these spectral properties would have allowed for a direct connection to the robustness analysis in Section \ref{sec_simplified_model}, potentially explaining \textit{which} specific error modes are dampened by the displacement convexity of the probability flow.
	
	\section{Conclusion}
	
	This paper presented a novel perspective on Score-based Generative Models by formulating the score estimation error as a stochastic source within a forward SPDE. This framework allows us to interpret the robustness of SGMs through the lens of geometric stability and displacement convexity. Based on this theory, we introduced SIEM, a metric derived from the quadratic variation of the SPDE solution.
	
	Our experiments demonstrate that SIEM serves as a robust proxy for sample quality, capturing the structural volatility of the generation process even with partial sampling. While further research is needed to explicitly characterize the spectral properties of the score error, this work highlights the potential of SPDE theory to provide both rigorous explanations for generative robustness and practical tools for efficient model evaluation. Future work should focus on identifying the specific eigenbasis of the error terms to bridge the gap between abstract stability bounds and the empirical behavior of neural networks.
	% ==================================================================
	% REFERENCES
	% ==================================================================
	
	% Use a .bib file in practice: 
    \bibliographystyle{unsrt}
	
	% Manual bibliography for demonstration:
	\bibliography{arxiv}
	%%%%%%%%%%%%%%%%%%%%%%%%%%%%%%%%%%%%%%%%%%%%%%%%%%%%%%%%%%%%%%%%%%%%%%%%%%%%%%%
%%%%%%%%%%%%%%%%%%%%%%%%%%%%%%%%%%%%%%%%%%%%%%%%%%%%%%%%%%%%%%%%%%%%%%%%%%%%%%%
% APPENDIX
%%%%%%%%%%%%%%%%%%%%%%%%%%%%%%%%%%%%%%%%%%%%%%%%%%%%%%%%%%%%%%%%%%%%%%%%%%%%%%%
%%%%%%%%%%%%%%%%%%%%%%%%%%%%%%%%%%%%%%%%%%%%%%%%%%%%%%%%%%%%%%%%%%%%%%%%%%%%%%%
\newpage
\appendix
\section{Adaptation of Denoising Score Matching for Tractable Estimation}
\label{DenoisingScoreMatchingForSIEM}
We give a tractable estimation in this section for the following $\xi_t$.
\begin{align}
	\xi(t):=d\mathbb{E}_{\mathbf{x}_t\sim v_{t}}\bigg[\bar{g}^{2}_{t}\frac{u_t}{v_t} (\mathbf{s}_{\boldsymbol{\theta}}(\mathbf{x}_t,t)-\nabla\log v_{t}(\mathbf{x}_t))\cdot\frac{\mathbf{x}_t}{\epsilon+\|\mathbf{x}_t\|^2} \bigg].
\end{align}
Recall that $v_t$ follows the reverse-time dynamics starting from $v_0:=p_T$. This implies that $v_t$ corresponds to the marginal distribution of the forward process at physical time $\tau = T-t$, i.e., $v_t = p_{T-t}$. To avoid confusion between the reverse time index $t$ and the forward process evolution, we introduce the forward time variable $\tau = T-t$.

The score model is based on the SDE $\ud\mathbf{x} = -f(\tau)\mathbf{x} \ud\tau + g(\tau)\ud\mathbf{w}$~\cite{song2021scorebased}. The transition kernel from time $0$ to $\tau$ is given by
\begin{align*} {p}(\mathbf{x}_\tau, \tau |\mathbf{y}, 0)=\mathcal{N}\left(\mathbf{x}_\tau; a(\tau,0)\mathbf{y}, b(\tau,0)^2 I \right) \end{align*}
where $a(\tau,0)$ and $b(\tau,0)$ are the standard coefficients derived from the forward SDE.
It is important to note that the integration variable $\mathbf{x}_t$ in $\xi_t$ is distributed according to $v_t$, which is equivalent to $\mathbf{x}_\tau$ sampled from $p_\tau$. Thus, $\mathbf{x}_t$ can be expressed by sampling noise $\mathbf{z} \sim \mathcal{N}(0,I)$ as follows:
\begin{align} \mathbf{x}_t = a(\tau,0)\mathbf{x}_0 + b(\tau,0)\mathbf{z}, \quad \text{where } \tau = T-t. \label{eq:xt_sampling} \end{align}

To adapt the expression, we employ a DSM-like technique~\cite{vincent2011connection}. The core idea is that the score model $\mathbf{s}_{\boldsymbol{\theta}}(\mathbf{x}_t, t)$ is trained to approximate the score of the conditional distribution $p(\mathbf{x}_t, \tau | \mathbf{x}_0, 0)$ rather than the marginal $v_t(\mathbf{x}_t)$. Therefore, we replace the marginal score $\nabla\log v_t(\mathbf{x}_t)$ with the conditional score $\nabla_{\mathbf{x}_t}\log p(\mathbf{x}_t, \tau | \mathbf{x}_0, 0)$.

The conditional score is derived as:
\begin{align*} \nabla_{\mathbf{x}_t} \log p(\mathbf{x}_t, \tau | \mathbf{x}_0, 0) &= \nabla_{\mathbf{x}_t} \left( -\frac{\|\mathbf{x}_t - a(\tau,0)\mathbf{x}_0\|^2}{2b(\tau,0)^2} \right) \\ &= -\frac{\mathbf{x}_t - a(\tau,0)\mathbf{x}_0}{b(\tau,0)^2} = -\frac{\mathbf{z}}{b(\tau,0)}. \end{align*}

Substituting this into the expression for $\xi_t$, the expectation $\mathbb{E}_{\mathbf{x}_t \sim v_t}$ is replaced by $\mathbb{E}_{\mathbf{x}_0 \sim p_0, \mathbf{z} \sim \mathcal{N}(0,I)}$. The term $(\mathbf{s}_{\boldsymbol{\theta}}(\mathbf{x}_t,\tau) -\nabla\log v_t(\mathbf{x}_t))$ becomes $\left(\mathbf{s}_{\boldsymbol{\theta}}(\mathbf{x}_t,\tau) + \frac{\mathbf{z}}{b(\tau,0)}\right)$.

Thus, the expression for $\xi_t$ transforms to:
\begin{align*} \xi(t) = \bar{g}^{2}_{t}\mathbb{E}_{\mathbf{x}_0 \sim p_0, \mathbf{z} \sim \mathcal{N}(0,I)}\bigg[\frac{u_t}{v_t}\left(\mathbf{s}_{\boldsymbol{\theta}}(\mathbf{x}_t,\tau) + \frac{\mathbf{z}}{b(\tau,0)}\right)\cdot \frac{\mathbf{x}_t}{\epsilon+\|\mathbf{x}_t\|^2} \bigg], \end{align*}
where $\mathbf{x}_t$ is determined by Eq. \eqref{eq:xt_sampling} with $\tau=T-t$. Note that $v_t$ in the denominator represents the true density $p_{T-t}(\mathbf{x}_t)$.
If we define the scaled score $\mathbf{s}_{\boldsymbol{\theta}}'$ as $-b(\tau,0)\mathbf{s}_{\boldsymbol{\theta}}$, we obtain:
\begin{align*} \xi(t) = \bar{g}^{2}_{t}\mathbb{E}_{\mathbf{x}_0 \sim p_0, \mathbf{z} \sim \mathcal{N}(0,I)}\bigg[-\frac{u_t}{v_t}\frac{1}{b(\tau,0)}\left(\mathbf{s}_{\boldsymbol{\theta}}'(\mathbf{x}_t,t) - \mathbf{z}\right)\cdot \frac{\mathbf{x}_t}{\epsilon+\|\mathbf{x}_t\|^2}\bigg]. \end{align*}

\section{Experimental Details}
\label{sec:appendix_experimental_details}

To ensure the reproducibility of our experiments and to provide a comprehensive account of our setup, this section details the configurations used for datasets, model architecture, training, and evaluation, as discussed in Section \ref{chap:experiments}.

\subsection{Datasets and Preprocessing}

\paragraph{Dataset Details.}
\begin{itemize}
	\item \textbf{CIFAR-10:} We used the standard training set of 50,000 images without any modifications to the official splits. The data was loaded from the \texttt{uoft-cs/cifar10} Hugging Face repository.
	\item \textbf{LSUN-Bedrooms:} For the LSUN-Bedrooms dataset, we used the full training set provided by the \texttt{pcuenq/lsun-bedrooms} Hugging Face repository.
\end{itemize}

\paragraph{Preprocessing.}
\begin{itemize}
	\item \textbf{Resizing:} The LSUN-Bedrooms images were resized to 32x32 pixels. The CIFAR-10 images were used at their original 32x32 resolution without resizing.
	\item \textbf{Normalization:} As mentioned in the main text, pixel values were normalized to the range [-1, 1]. This was achieved by first converting images to Tensors (scaling to [0, 1]) and then applying a normalization with mean=0.5 and standard deviation=0.5.
\end{itemize}

\paragraph{Data Augmentation.}
\begin{itemize}
	\item We did not use data augmentation during training. The code includes commented-out lines for random horizontal flips, but they were not active in the final configuration.
\end{itemize}

\subsection{Model Architecture Details}
The experiments utilized a U-Net architecture based on the \texttt{diffusers} library's \texttt{UNet2DModel}. The specific configuration is detailed below.

\paragraph{U-Net Hyperparameters.}
\begin{itemize}
	\item \textbf{Base channel count:} 128
	\item \textbf{Channel multipliers:} The \texttt{block\_out\_channels} were explicitly set to (128, 256, 256, 256). This corresponds to channel multipliers of (1, 2, 2, 2) relative to the base channel count for resolutions 32x32, 16x16, 8x8, and 4x4, respectively.
	\item \textbf{Residual blocks per resolution:} 2
	\item \textbf{Attention resolutions:} Self-attention mechanisms were applied at the 16x16 resolution, as indicated by the \texttt{AttnDownBlock2D} and \texttt{AttnUpBlock2D} blocks at the second level of the U-Net.
\end{itemize}

\paragraph{Components.}
\begin{itemize}
	\item \textbf{Normalization:} Group Normalization was used throughout the network (default for this model).
	\item \textbf{Activation function:} SiLU (also known as Swish) was used as the non-linear activation function (default for this model).
	\item \textbf{Time embedding:} Time was encoded using a sinusoidal positional embedding, which was then processed by a small MLP before being added to the residual blocks (default for this model).
\end{itemize}

\subsection{Training and Diffusion Configuration}

\paragraph{Optimization.}
\begin{itemize}
	\item \textbf{Optimizer:} AdamW
	\item \textbf{Learning rate:} A learning rate of 1e-4 was used with a cosine learning rate schedule, preceded by a linear warm-up phase of 500 steps.
	\item \textbf{Batch size:} 128
	\item \textbf{Other parameters:} Default AdamW parameters were used ($\beta_1=0.9$, $\beta_2=0.999$). Gradient clipping was applied with a maximum norm of 1.0.
\end{itemize}

\paragraph{Exponential Moving Average (EMA).}
\begin{itemize}
	\item An exponential moving average (EMA) of the model weights was not used in the provided training script.
\end{itemize}

\paragraph{Diffusion Configuration.}
\begin{itemize}
	\item \textbf{Noise schedule:} A linear noise schedule was used over 1000 diffusion timesteps, as implemented by the default \texttt{DDPMScheduler} from the \texttt{diffusers} library.
	\item \textbf{Schedule parameters:} The default \texttt{diffusers} library parameters were used: $\beta_{start}$ was set to 1e-4 and $\beta_{end}$ was set to 0.02.
\end{itemize}

\subsection{Evaluation Details}

\paragraph{FID (Fréchet Inception Distance).}
\begin{itemize}
	\item \textbf{Implementation:} The FID score was computed using the \texttt{FrechetInceptionDistance} metric from the \texttt{torchmetrics} library.
	\item \textbf{Sample Count:} The evaluation was performed using 10,000 real images from the test set and 10,000 generated images.
	\item \textbf{Image Preprocessing:} For the FID calculation, both real and generated images were transformed into tensors with pixel values scaled to the range [0, 1], as required by the metric implementation. Normalization to [-1, 1] was not applied for this specific evaluation.
	\item \textbf{Image Generation:} Generated samples were produced using the \texttt{diffusers.DDPMPipeline} with the corresponding trained U-Net model and the default DDPM scheduler, which uses 1000 inference steps. Generation was performed in batches of 5,000.
	\item \textbf{Statistics:} The final reported FID score is the mean and standard deviation calculated over 5 independent evaluation runs.
\end{itemize}

\paragraph{2-Wasserstein Distance.}
\begin{itemize}
	\item \textbf{Implementation:} The 2-Wasserstein distance was estimated using the Sinkhorn algorithm provided by the \texttt{geomloss} library, specifically via its \texttt{SamplesLoss} function.
	\item \textbf{Sinkhorn Parameters:} The Sinkhorn approximation was configured with $p=2$, a regularization parameter \texttt{blur} of 0.05, and a \texttt{scaling} parameter of 0.9.
	\item \textbf{Sample Count:} The distance was computed between 10,000 real images from the test set and 10,000 generated images.
	\item \textbf{Image Preprocessing:} Both real and generated images were normalized to the range [-1, 1] and then flattened into vectors before being passed to the distance function.
	\item \textbf{Image Generation:} Generated samples were produced by running the full 1000-step DDPM reverse process from random noise.
	\item \textbf{Statistics:} The final reported 2-Wasserstein distance is the mean and standard deviation calculated over 5 independent evaluation runs.
\end{itemize}

\paragraph{SIEM (SPDE-Induced Evaluation Metric)}
\begin{itemize}
	\item \textbf{Implementation:} The metric was computed by iterating through the reverse diffusion process. A key challenge is estimating the density ratio $\frac{u_t}{v_t}$ at each timestep $t$. We addressed this by training a small, auxiliary binary classifier—a Convolutional Neural Network (`DensityRatioClassifier`)—at each evaluated timestep. This classifier was trained to distinguish between generated samples at that timestep ($X_k$) and real data noised to the same level. The output of the trained classifier was then used to estimate the required density ratio.
	\item \textbf{Classifier Training Details:} The auxiliary classifier was trained using the Adam optimizer with a learning rate of 1e-3 and a BCEWithLogitsLoss function. Training was conducted for up to 50 epochs with a batch size of 512, utilizing early stopping with a patience of 5 based on a 20\% validation split. The classifier's weights were re-initialized at each diffusion timestep to ensure an unbiased estimation for that specific time.
	\item \textbf{SIEM Calculation Parameters:} The metric was computed using 10,000 samples. To stabilize the density ratio estimation, the estimated values were clamped to the range [0.01, 100.0]. The evaluation was performed in two settings: using the full 1000 reverse diffusion timesteps, and using only the final 100 timesteps (from 999 down to 900).
	\item \textbf{Component Separation:} The final SIEM score is the square root of the mean of the squared values computed at each timestep, as per Equation \ref{SIEM}. For the correlation analysis presented in Table \ref{tab:correlation_matrix}, the systematic bias component, $\boldsymbol{\mu}_t$, was estimated by applying a 1D Gaussian filter (with sigma=10) to the time series of computed values. The SIEM component in the tables refers to the L2 norm of the residual (the original time series minus the smoothed bias).
	\item \textbf{Statistics:} The final reported SIEM score is the mean and standard deviation calculated over 5 independent runs.
\end{itemize}	
\end{document}